\pdfoutput=1

\documentclass[11pt]{article}

\usepackage[preprint]{acl}

\usepackage{subfig}
\usepackage{times}
\usepackage{latexsym}
\usepackage{amsthm,amsmath,amssymb}
\usepackage{graphicx}

\usepackage{mathrsfs}
\usepackage{url}
\usepackage{float}
\usepackage{diagbox}

\usepackage[T1]{fontenc}

\usepackage[utf8]{inputenc}

\usepackage{microtype}

\usepackage{inconsolata}
\usepackage{color}
\usepackage{units}
\usepackage{multirow}
\usepackage{tabularx}
\usepackage{hyperref}

\usepackage{graphicx}
\usepackage{makecell}

\usepackage{soul}

\title{SimCT: A Simple Consistency Test Protocol  in LLMs Development Lifecycle}

\author{ Fufangchen Zhao$^{+*}$ \textsuperscript{\rm1,\rm2,\rm3}
    Guoqiang Jin$^{*}$ \textsuperscript{\rm3}
    Jiaheng Huang \textsuperscript{\rm3}
    Rui Zhao \textsuperscript{\rm3}
    Fei Tan$^{\dagger}$ \textsuperscript{\rm3}
    \\
  \textsuperscript{1} Beijing University of Posts and Telecommunications,  Beijing, China \\
  \textsuperscript{2} State Key Laboratory of Networking and Switching Technology \\
  \textsuperscript{3} SenseTime Research \\
  \texttt{zhaofufangchen@bupt.edu.cn} \\
  \texttt{\{jingguoqiang, huangjiaheng, zhaorui, tanfei\}@sensetime.com}
  }

\begin{document}
\maketitle

\def\thefootnote{+}\footnotetext{Work was done during internship at SenseTime Research}
\def\thefootnote{*}\footnotetext{Equal contribution}
\def\thefootnote{$\dagger$}\footnotetext{Corresponding author}

\begin{abstract}

In this work, we report our efforts to advance the standard operation procedure of developing Large Language Models (LLMs) or LLMs-based systems or services in industry. We introduce the concept of Large Language Model Development Lifecycle (LDLC) and then highlight the importance of consistency test in ensuring the delivery quality. The principled solution of consistency test, however, is usually overlooked by industrial practitioners and not urgent in academia, and current practical solutions are insufficiently rigours and labor-intensive. We thus propose a simple yet effective consistency test protocol, named SimCT. SimCT is mainly to proactively check the consistency across different development stages of "bare metal" LLMs or associated services without accessing the model artifacts, in an attempt to expedite the delivery by reducing the back-and-forth alignment communications among multiple teams involved in different development stages. 

Specifically, SimCT encompasses response-wise and model-wise tests. We implement the protocol with LightGBM and Student's t-test  for two components respectively, and perform extensive experiments to substantiate the effectiveness of SimCT and the involved components. Our code and protocol demo are available at \url{https://github.com/SenseTime-Research-Language/SimCT}

\end{abstract}

\section{Introduction}
\label{intro}

\begin{figure}[t]
   \centering
   \includegraphics[width=\linewidth]{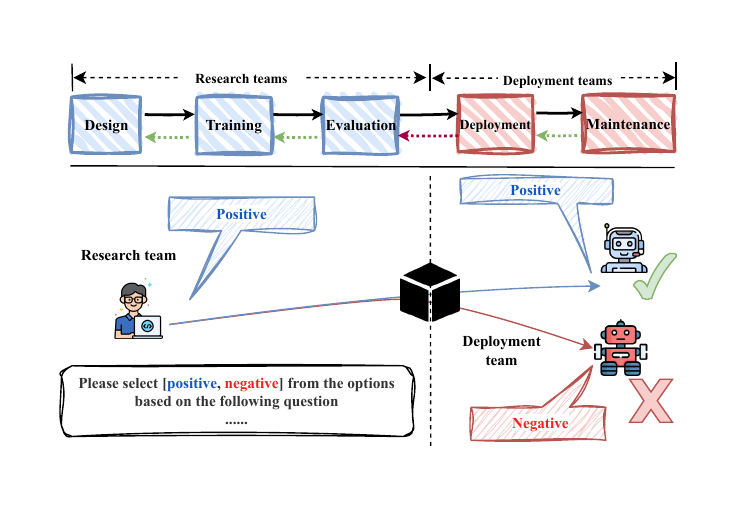}
    \caption{Diagram of LDLC and an example illustrating the necessity of consistency test for LLM-based systems in typical development stages. }  
    \label{consistency}
    \vspace{-5pt}
\end{figure}

Large language models (LLMs) have revolutionized natural language processing (NLP) with their significantly improved text generation capabilities \citep{jakesch2023human,sadasivan2023can, lu2022makes,zhang2024balancing}, promoting the rapid development of NLP. The GPT family \citep{ouyang2022training}, launched by OpenAI, which includes the popular ChatGPT and GPT-4, has become one of the most widely used LLMs in industrial applications. Concurrently, the open-source community has also given rise to non-commercial LLMs such as Llama \citep{touvron2023llama}, GLM \citep{zeng2022glm}, and Mistral \citep{jiang2023mistral}, contributing to technological diversity. Furthermore, LLM-based technologies, such as RAG (Retrieval-Augmented Generation) \cite{gao2023retrieval} and Agent \cite{bubeck2023sparks} applications, are being increasingly widely adopted, further solidifying LLMs' leadership in the field of NLP.


In software development, teams commonly employ the Software Development Lifecycle (SDLC) methodology to design and construct high-quality software \citep{Praise2022sdlc}. This cost-effective and time-efficient process aims to reduce project risks through proactive planning, ensuring that the software meets customer expectations both during production and in the long term. Likewisely, in industrial LLMs research and development projects, teams are expected to adhere to a comparable procedure which we refer to as the Large Language Model Development Life Cycle (LDLC) as delineated in Fig. \ref{consistency}.


In industrial scenarios, multiple teams  usually collaboratively work on different development stages of LLMs. Consistency testing is essential to guarantee that these teams consistently deliver model or system artifacts. Unfortunately, LDLC is yet to be studied or executed fully as opposed to SDLC. Consistency test involved in LDLC, however, is urgent and challenging. 
Multiple factors may shape the consistency checking. Firstly, the decoder structure of LLMs intrinsically introduces randomness amid the generation. Secondly, external factors include differentiation in hardware architectures (e.g., Nvidia's Ampere/Volta and TPU), heterogeneous inference and deployment tools among multiple teams throughout the development course. Moreover, misaligned hyper-parameters (e.g., temperature coefficient), may also lead to different responses systematically.
The interplay between intrinsic and external factors complicates the task of ensuring the consistency of LDLC deliverables, as illustrated in Figure \ref{consistency}.

In this context, we propose SimCT, a \underline{Sim}ple \underline{C}onsistency \underline{T}est protocol in LDLC. It attempts to pinpoint impacts of external factors (e.g., devices, hyper-parameters) on LLMs or LLMs-centric systems conditioned on the responses or outputs for proactively diagnosing potential gaps throughout different development stages. Specifically, SimCT involves response-wise and model-wise tests as detailed in Section \ref{SimCT_protocol}.
We formulate the former as a binary classification, which takes as input features classical generation metrics (e.g., BLEU \citep{papineni2002bleu}, ROUGE \citep{lin2004rouge}, METEOR \citep{banerjee2005meteor}) to determine if two responses are generated by the same model. We then aggregate response pairs of all queries for two models and perform model-wise consistency test utilizing statistical t-test. To facilitate the development and evaluation of SimCT, we have also crafted consistency test datasets and benchmarks in this regard.

Our contributions are summarized as: 1) We introduce the concept of LDLC to describe the current industrial procedure of large model development; (2)We emphasize the essential role of consistency test in LDLC to facilitate the synergy across different teams; (3) We propose SimCT protocol and craft a pertinent dataset to instantalize consistency test and demonstrate its effectiveness and generalization.

\section{Related Work}
\begin{figure*}[h]
   \centering
   \includegraphics[width=\linewidth]
    {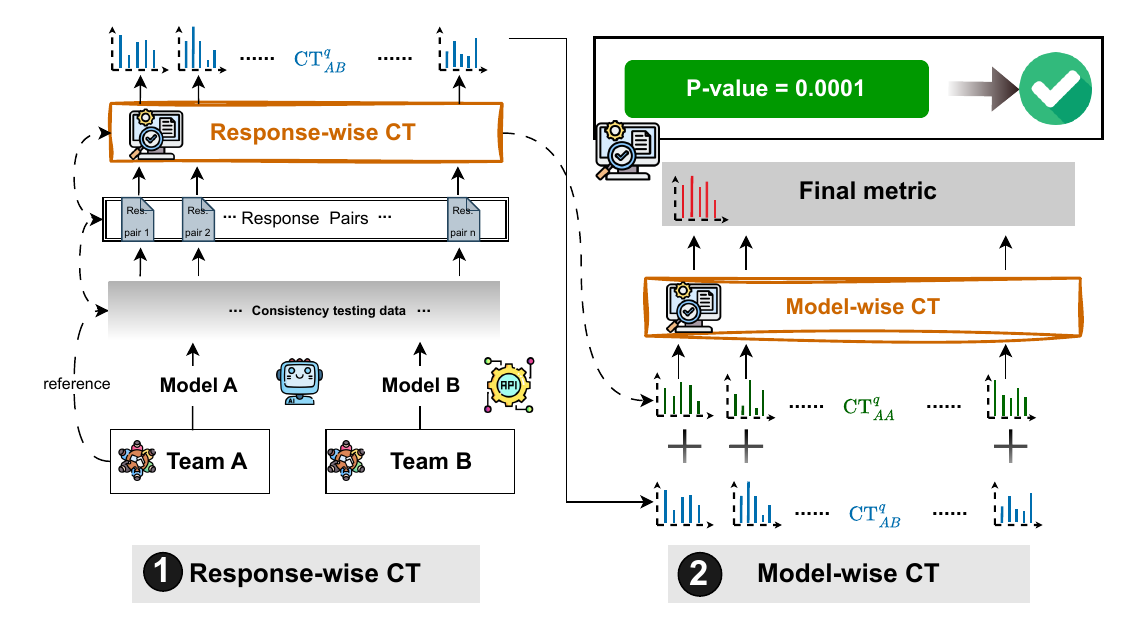}
    \caption{The overall framework of SimCT involves the response-wise test and the model-wise test (aggregating all response pairs).
    }
    \label{SimCT}
\end{figure*}
\subsection{Model Generation Analysis}
Previous research on single model generation often employs traditional evaluation metrics to quantitatively assess the model-generated content. BLEU \citep{papineni2002bleu} is currently the predominant metric for evaluating machine translation, whereas ROUGE \citep{lin2004rouge} assesses generation performance based on recall rate. METEOR \citep{banerjee2005meteor}, an enhancement of BLEU, offers a more user-friendly approach. These metrics, however, have been criticized for their inability to effectively reflect the NLG system's performance \citep{stent2005evaluating,reiter2009investigation} especially in LLMs era. Recently, GPTScore \citep{fu2023gptscore} utilizes LLMs (e.g., GPT-3) for evaluation purposes. \citet{liu2023gpteval} has suggested employing GPT-4 as a powerful evaluator to achieve better alignment with human judgment. They aim to evaluate the quality of generated content based on human intent. Our work, however, capitalize on the existing metrics to address the consistency issue among multiple parties.


Beyond this, multi-model analysis in generation is currently limited, yet there is still some relevant work. LMdiff \citep{strobelt2021lmdiff} assesses differences between models by conducting white-box analysis. Chatbot Arena \citep{zheng2023judging} enables users to compare the responses of two models through a black-box approach and rank them accordingly.
However, the automatic consistency analysis tool in a black-box manner is yet to be explored.

\subsection{LLM Consistency}
There are also some studies on the consistency of large language models (LLMs). \citet{raj2022measuring} measures the reliability of LLMs through semantic consistency evaluations. \citet{tam-etal-2023-evaluating}  assesses the factual consistency of LLMs, specifically in the context of news summarization. \citet{cui2023evaluating} proposes DCR, an automated framework that employs a divide-and-conquer reasoning strategy to evaluate and enhance the consistency of LLM-generated text against human judgment. \citet{raj2023semantic} underscores the significance of fully understanding LLM consistency in open text generation scenarios. \citet{chen2023universal} introduces the concept of universal self-consistency aimed at improving the performance of LLM generation.
The consistency here mainly refers to whether models/systems are consistent with the ground truth or human preference. Our work, however, emphasizes if models developed at different stages are intricately consistent regardless of they align with ground truth or human preference as well as develops a consistency test protocol. It is an essential effort to advance the standard operation procedure for LLMs development in industry. By introducing automatic end-to-end consistency testing into LDLC, we aim to ensure the delivered services are consistent with what LLMs are developed for initially.

\section{SimCT Protocol}
\label{SimCT_protocol}
The pivotal parts of SimCT are to perform response-wise consistency test and then predict the model-wise consistency by evaluating and aggregating typical response pairs from different models. The overall framework of SimCT is delineated in Figure \ref{SimCT}.

\subsection{Response-wise CT}
To determine whether models $A$ and $B$ are consistent with each other, we can only pinpoint relevant signals from their responses if model parameters are not accessible. Specifically, given query $q$, we have two responses $r_A^q$ and $r_B^q$ from models $A$ and $B$, respectively. The challenge here is that two identical models (e.g., calling the same API twice) don't necessarily give the exactly same responses due to sampling nature, and two different models (e.g., models of different parameter scale) are likely to respond similarly given some query. We thus introduce classical metrics including ROUGE, BLUE, METEOR, and DENSE\footnote{We use m3e \cite{m3e} embedding model to generate  semantic similarity as dense features.} to measure
the consistency signals of response pairs at the lexical/semantic level. Furthermore, the TYPE of the query strongly influences the consistency of the responses, so we roughly divide them into two categories, i.e., closed-end and open-end. Formally, we have response-wise consistency test: 
\begin{equation}
\begin{aligned}
\label{form_r_ct}
\mathrm{CT}_{\mathrm{AB}}^{\mathrm{q}}= & f_R\left(\operatorname{ROUGE}\left(r_A^q, r_B^q\right), \operatorname{BLEU}\left(r_A^q, r_B^q\right),\right. \\
& \operatorname{METEOR}\left(r_A^q, r_B^q\right), \operatorname{DENSE}\left(r_A^q, r_B^q\right), \\
& \operatorname{TYPE}(q))
\end{aligned}
\end{equation}
where $f_R$ is a model-agnostic learning function to map the query and the associated responses into consistency likelihood.



Intuitively, it is not statistically significant to evaluate the consistency of LLMs or associated system quantitatively based on the response pair of a single query. Furthermore, it's challenging to pinpoint the consistent signals by ruling out intrinsic randomness due to sampled decoding as mentioned in Section \ref{intro}. Therefore, it's demanding to apply response-wise consistency test to more queries for amplifying the patterns.

\subsection{Model-wise CT}
\label{Model-wiseCT}
In this context, we propose model-wise consistency
test, which is the end-to-end goal of simCT as well. 
Specifically, given models $A$ (assumed as the upstream one) and $B$ (assumed as the downstream one) and query set $Q = {q_1, q_2, \ldots, q_N}$, the protocol has the following steps:
\begin{itemize}
    \item Given query $q_i$, collect two sampled responses $r_A^{q_i}$ and $\bar{r}_A^{q_i}$ from model $A$, and single response $r_B^{q_i}$ 
    \item Utilize Eq. \ref{form_r_ct} to estimate response-wise consistency likelihood $\mathrm{CT}_{AB}^{q_i}$ for two models and $\mathrm{CT}_{AA}^{q_i}$ as response-pair 
    reference of upstream model $A$
    \item Armed with set $\{\mathrm{CT}_{AB}^{q_i}, \mathrm{CT}_{AA}^{q_i}\}_{i=1}^N$, we perform classical Student's t-test for paired samples with null hypothesis that the mean of $\mathrm{CT}_{AB}^{q_i}$ is not equal to the mean of $\mathrm{CT}_{AA}^{q_i}$. If the test is statistically significant (e.g., p-value <= 0.05), models A and B are regarded as consistent, otherwise it's inconsistent. 
\end{itemize}
The comparison with reference helps to get rid of intrinsic randomness involved in the decoding.


\subsection{Dataset Craft}

It's noted that $f_R$ involved in the proposed SimCT 
is supposed to be learnt from model responses and generalized 
in unseen scenarios. Therefore, we report the way we craft training and test/benchmark datasets to simulate the real-world applications as detailed in 
Fig. \ref{data} in Appendix.

\subsubsection{Training dataset}
\label{train}
 We follow InstructGPT \citep{ouyang2022training} to 
 gather passages from Chinese Wikipedia\footnote{\url{Wikipedia.org}} and utilize ChatGPT to generate corresponding questions for them. Following a manual screening procedure to evaluate quality and difficulty meticulously, we obtain 1,000 high-quality, medium-difficulty queries. As indicated in Eq. \ref{form_r_ct},  we categorize the queries in our dataset into `open-end' and `closed-end' types, encoded as $1$ and $0$, respectively.
 They are roughly 50-50 split (52.1\% vs 47.9\%) to simplify the model training without loss of generality. 

In this work, consistent model pairs refer to the final decoding distribution remains unchanged given all queries. That's to say, all possible factors don't alter decoding distribution throughout the development lifecycle. It's inconsistent if the distribution shifts due to any intrinsic or external factors as mentioned before.

Specifically, given 1,000 queries, we have 4 different LLMs as detailed in Table \ref{models} in Appendix to build 7 consistent and 8 inconsistent model pairs, respectively. 
For consistent model pairs, we generate 6,263 response pairs by calling the same LLMs or APIs twice. For inconsistent model pairs, we generate 7,518 response pairs by simulating the possible scenarios (e.g., different LLMs or the same LLMs with different hyper-parameters or deployment environment) as detailed in Appendix \ref{appendix:Dataset Construction Rules}.
In total, we have 13,781 response-wise training samples to develop $f_R$.







\subsubsection{Test Dataset}
\label{test dataset}
Regarding the test set, we collect 
138 high-quality queries with 48.6\% and 51.4\% being open-end and closed-end respectively. As with training part, we build 14 consistent model pairs and 
15 inconsistent model pairs as described in Table \ref{models} in Appendix. We thus have a total of 4,002 response pairs. In addition, to facilitate the model-wise CT as mentioned before, we generate references for 4,002 response pairs by calling the upstream model one more time.


\subsection{Implementation Instance}
For $f_R$ in response-wise test, we introduce LightGBM \citep{ke2017lightgbm} as the classifier due to its lightweight serving and compelling capability.
The specific implementation details (such as parameter settings) are shown in Table \ref{param} in Appendix.

Regarding model-wise test, we have the null hypothesis that the mean of $\mathrm{CT}_{AB}^{q_i}$ is not equal to the mean of $\mathrm{CT}_{AA}^{q_i}$, if the test is statistically significant (i.e., p-value <= 0.05), models A and B are regarded as consistent. It's inconsistent if it is not statistically significant (i.e., p-value >= 0.95)\footnote{It's noted that p-value >= 0.95 is equivalent to that the null hypothesis of their means are equal with p-value <= 0.05.}.
For cases without statistical significance (i.e., 0.05 < p-value < 0.95), we regard them as inconsistent. 
The rationale is that if it's falsely predicted as consistent, the potential impact is worse than the other way around in industrial practice and leads to more workloads.


\section{Experiment } 
\subsection{Baselines}
\label{baseline}

To make the entire consistency testing experiment more robust and comprehensive, we introduce the following two different baselines:

For the first baseline, we select GPT-4o as the single-stage test scheme by using all the response pairs $\{r_A^{q_i}, r_B^{q_i}\}^N_{i=1}$ straightly from model pair (mentioned in Section \ref{test dataset}) as input. This allow GPT-4o to directly determine whether the model pair is consistent. To avoid randomness, we replicate the experiment 10 times for each model pair. We then calculate the overall judgement accuracy as the baseline metric. 

For the second baselines (dual-stage schemes), we instantiate $f_R$ with advanced models such as RoFormer \citep{su2024roformer}, GLM4-9B \citep{zeng2022glm,du2022glm,wang2023cogvlm} and GPT-4o to derive the response-wise $\mathrm{CT}^q_{AB}$ for the given query $q$. As vanilla RoFormer cannot generate reasonable quantitative metrics  with only prompts like GLM4-9B and GPT-4o, we additionally fine-tune RoFormer on the training set (Section \ref{train}). 
 For human annotation,  we recruit seven graduate students in NLP/LLMs to  independently vote the consistency of all response pairs. We take the ratio of voting for consistency as response-wise score (e.g., 5/7). We then employ a {\it Threshold} (majority voting) as described in Appendix \ref{evaluation metrics} and {\it T-test} to ascertain whether it's consistent or inconsistent for all methods.


\subsection{Main Result}
As shown in Table \ref{results}, our SimCT is significantly superior to other baselines.  Especially, SimCT surpasses the best setting of GPT-4o, which substantiates the necessity of SimCT dedicated to consistency test.
It's also noted that human annotation doesn't necessarily excel in diagnosing the consistency among models, further suggesting the complication of this task.

\begin{table}[t]
    \Large
    \centering
    \renewcommand\arraystretch{1.5}
     \scalebox{0.47}{
    \begin{tabular}{cccccc}
    \hline
          & GPT-4o & Human  & RoFormer & GLM4-9B & SimCT\\ 
          \hline
          Single-stage &68.97 & -- & -- & -- & -- \\
          \hline
          Threshold (dual-stage) & 65.52 & 61.72 & 63.45 & 55.17 & \textbf{79.31}\\
          \hline
          T-test (dual-stage) & 86.21 & 62.07 & 82.76 & 75.86 & \textbf{93.10}\\
          \hline
          Optimal  & 86.21 & 62.07 & 82.76 & 75.86 & \textbf{93.10} \\
          
          \hline
    \end{tabular}
    }
    \caption{Accuracy comparison of our SimCT and diverse baselines. `--' stands for no experiments performed.}
    \label{results}
\end{table}



Furthermore,  SimCT with LightGBM being response-wise $f_R$ performs significantly better than other baselines across both Threshold and T-test schemes. T-test as model-wise protocol in the two-stage scheme is also generally better than threshold protocol, substantiating its reasonability and statistical rationality.

\subsection{Ablation Study}

\begin{table}[t]
    \Large
    \centering
    \renewcommand\arraystretch{1.5}
     \scalebox{0.47}{
    \begin{tabular}{cccccc}
    \hline
          & GPT-4o & Human  & RoFormer & GLM4-9B & SimCT\\ 
          \hline
          Optimal &86.21 & 62.07 & 82.76 & 75.86 &\textbf{93.10}\\
          \hline
          
          Optimal (w/o TYPE)  &72.21 & 55.86 &57.74 & 62.04 & \textbf{82.76}\\
          \hline
    \end{tabular}
    }
    \caption{Comparison of optimal results and their counterparts without question type (TYPE). For GPT-4o, question type is taken off from the prompt.}
    \label{query type}
\end{table}

As presented in Table \ref{query type}, the query type largely shapes the consistency test performance across our proposed SimCT and baselines. 
Intuitively, 
if two different students of similar level are asked to answer a multiple-choice question (closed-end) and write an email (open-end), the responses of the latter can provide more differential signals. likewisely, if the case is to ask the same students twice, multiple-choice question (closed-end) is more helpful.

\subsection{Case Study}
To facilitate the understanding of our simCT intuitively, we perform response-wise and model-wise case studies.

\subsubsection{Response-wise}
\begin{table*}[t]
    \centering
    \renewcommand\arraystretch{2}
     \scalebox{0.7}{
    \begin{tabular}{|c|c|c|c|c|}
    \hline
          Query & Response A & Response B &  SimCT (LightGBM) & RoFormer\\ 
          \hline
            \makecell[c]{What is the main food of the\\ Three Striped Flag Seabream?} & Coral. & \makecell[c]{The three banded flag snapper feeds \\ mainly on coral polyps.} & 56.76 & 58.64\\
          \hline
           \makecell[c]{Please select [positive, negative] from the \\ options based on the following question \\ $\dots$} & Positive & Positive & 88.54 & 73.31\\
          \hline
          \makecell[c]{What is the largest known organism at \\ present?} & \makecell[c]{The currently known tallest \\ plant is the African cotton tree.} &  \makecell[c]{The largest known organism \\ is the blue whale.} & 21.31 & 38.84\\
          \hline
          
    \end{tabular}
    }
    \caption{Three typical queries and their responses from different models checked by SimCT and RoFormer with reponse-wise test, where all SimCT values are multiplied by 100.}  
    
    \label{case study 2}
\end{table*}
We randomly sample three different queries from the test set, as shown in Table \ref{case study 2}.  It is evident that for identical response pairs, $f_R$ tends to give a very high score, while for responses that convey the same meaning but differ in content, $f_R$ gives a relatively medium score. For completely different response pairs, the score is quite low. Although the scores vary for different response pairs, it remains tricky to directly determine whether the underlying models are consistent. This intuitively highlights the rationality of our protocol of combining response-wise and model-wise tests.

\subsubsection{Model-wise}
\begin{table}[t]
    \Large
    \centering
    \renewcommand\arraystretch{1.5}
     \scalebox{0.55}{
    \begin{tabular}{|c|ccc|}
    \hline
         \diagbox{Model A}{Model B} & \makecell[c]{Baichuan-13B\\$\tau$=0.2} & \makecell[c]{Baichuan-7B\\$\tau$=0.2} & \makecell[c]{Baichuan-13B\\$\tau$=0.5}\\ 
          \hline
          \makecell[c]{Baichuan-13B\\$\tau$=0.2} &4.87e-17 &  0.99 &0.79  \\
          \hline
          \makecell[c]{Baichuan-7B\\$\tau$=0.2} &0.99 & 3.58e-10 & 0.99 \\
          \hline
          \makecell[c]{Baichuan-13B\\$\tau$=0.5} & 0.82 & 0.99 & 0.02\\ 
          \hline
    \end{tabular}
    }
    \caption{Illustrative cases of model-wise test.}
    \label{case study}
\end{table}

To further demonstrate the statistical significance of our SimCT, we sample the p-value results of the t-test for Baichuan-7B and Baichuan-13B \citep{baichuan2023baichuan2}, as detailed in Table \ref{case study}. For inconsistent model pairs (Baichuan-7B vs Baichuan-13B), the p-value is close to 1, while for consistent model pairs, the p-value is close to 0.
For model pairs differing in temperature coefficient (e.g., Baichuan-13B with $\tau = 0.2$ vs $\tau = 0.5$), the p-value is not that informative (e.g., p-value = 0.79). 
This further confirms the challenges of pinpointing 
the subtle variations involved in model development and deployment.

\section{Discussion}



LLMs are regarded as the cutting-edge technology and have gradually been employed in industry. The routine development work mainly consists of data cleaning, model training, performance evaluation and deployment. 
Performance evaluation is a highly frequent item though it's not as fancy as  model training throughout LDLC. The back and forth sync-up and double checking, however, is time-consuming. Unfortunately, the most typical solution is far from being principled.


Our SimCT protocol is developed to bridge this gap. It can not only improve the efficiency of large teams or multiple teams working together, but also effectively avoid the butterfly effect caused by small problems in the R\&D procedure and ensure the smooth progress of the project. In addition, our work has far-reaching implications and will inspire further exploration and research into safety and authenticity in large-scale model industrial R\&D in the future. This not only provides a solid foundation for existing technology, but also paves the way for future innovation. Our work highlights the importance of SOP in LLMs development and the involved test (e.g., the proposed consistency test) to ensure the SOP can be executed.




\section{Conclusion}

In this work, we develop SimCT, a simple yet principled consistency test protocol in LLMs development lifecycle, dedicated to addressing common consistency issues involved in the industrial R\&D procedure of LLMs. Furthermore, we have made the relevant code and datasets publicly available, detailing the construction procedure of these datasets to facilitate the community to further advance the SOP of LLMs development.

\section{Limitations}
There are inevitably differences between 
our experimental settings and the real application scenarios of LLMs development. 
We can't cover all scenarios though we try our best.

\bibliography{custom,consistency}

\newpage
\appendix


\section{Prompts for Quantitative Indicators}
\label{appendix:Zero-shot Prompt}
Our prompt is as follows:

\textit{Now, here are two paragraphs for you. These two texts are two outputs generated by two models. These two models may be the same or different. Please confirm if the two models used to generate these two texts are the same, and give a score between "0" and "1", where "0" represents that the two models are different and "1" represents that the two models are the same, with the result rounded to three decimal places.}

\textit{Requirement: 1) The score cannot be equal to "0" or "1"! 2) Provide the reason for giving this score!}

\textit{You can use the following indicators as a basis for judgment:}

\textit{1. The semantic similarity between two paragraphs of text. Generally, the semantic similarity between replies obtained from the same question is relatively high.}

\textit{2. Differences in narrative logic between the two texts. Generally, there are significant differences in the narrative logic of different models, while the narrative logic of the same model is basically the same.}

\textit{3. The generation quality of two paragraphs of text generally varies among different models, and the generation quality of the same model is basically similar.}

\textit{4. The generation confidence of two paragraphs of text is generally different for different models, while the generation confidence of the same model is basically the same. }

\begin{table*}[h]
    \centering
    \scalebox{1.0}{
    \begin{tabular}{|c|c|c|c|c|c|}
        \hline
         \textbf{Param}  & \textbf{Value} & \textbf{Param} & \textbf{Value} & \textbf{Param} & \textbf{Value}\\
        \hline
        objective  & binary & metric & auc & num\_leaves & 20 \\
         max\_bin & 40 & max\_depth & 2 & learning\_rate & 0.1 \\
         colsample\_bytree & 0.9 &  bagging\_fraction & 0.9 & min\_child\_samples & 1\\
         n\_jobs & -1 & silent & True & seed & 1 \\
         \hline
    \end{tabular}
    }
    \caption{Implementation details in LightGBM fo SimCT.}
    \label{param}
\end{table*}

\begin{table*}[h]
    \centering
    \scalebox{1.0}{
    \begin{tabular}{c|cc}
        \hline
         train  & \multicolumn{2}{c}{test} \\
        \hline
         Ziya \citet{fengshenbang}  &Ziya &SenseChat-$\alpha$ \\
         SenseChat-$\alpha$ \citep{sensetime2023} & SenseChat-$\beta$ &SenseChat-$\gamma$ \\
         SenseChat-$\beta$ & Baichuan2-7B \citet{baichuan2023baichuan2} & Bichuan-13B \\
         SenseChat-$\gamma$ & LLaMa-70B \citet{touvron2023llama} & Qwen \citet{qwen} \\
         & Llama2-Chinese-7B \citet{Chinese-LLaMA-Alpaca} &Llama2-Chinese-13B\\
         \hline
    \end{tabular}
    }
    \caption{Detailed LLMs involved in dataset crafting.}
    \label{models}
\end{table*}

\section{Threshold-based Test}
\label{evaluation metrics}
Given $\{\mathrm{CT}_{AB}^{q_i}\}^N_{i=1}$, the overall consistency likelihood $\mathrm{CT}_{AB}$ is as follows:
\begin{equation}
    \mathrm{CT}_{AB} = \frac{\sum\limits_{i=1}^N\left[\mathrm{CT}^{q_i}_{AB}\geq {\lambda_{\mathrm{response}}}\right]}{N}
\end{equation}
where $N$ represents the number of response pairs in a single case as mentioned in main text, and $\lambda_{\mathrm{response}}$ is a response-wise hyper-parameter optimized for different baselines. It should be noted that $[\cdot]$ denotes an Iverson bracket\footnote{$[\mathrm{P} ]= \begin{cases}1 & \text { If } P \text { is true } \\ 0 & \text { Otherwise }\end{cases}$} \citep{knuth1992two}. 
For model-wise $\mathrm{CT}_{AB}$, we take the majority voting (more than 1/2) to decide if the model is consistent or not.


\section{Dataset Construction Rules}
\label{appendix:Dataset Construction Rules}




\begin{figure}[t]
   \centering
   \includegraphics[width=\linewidth]{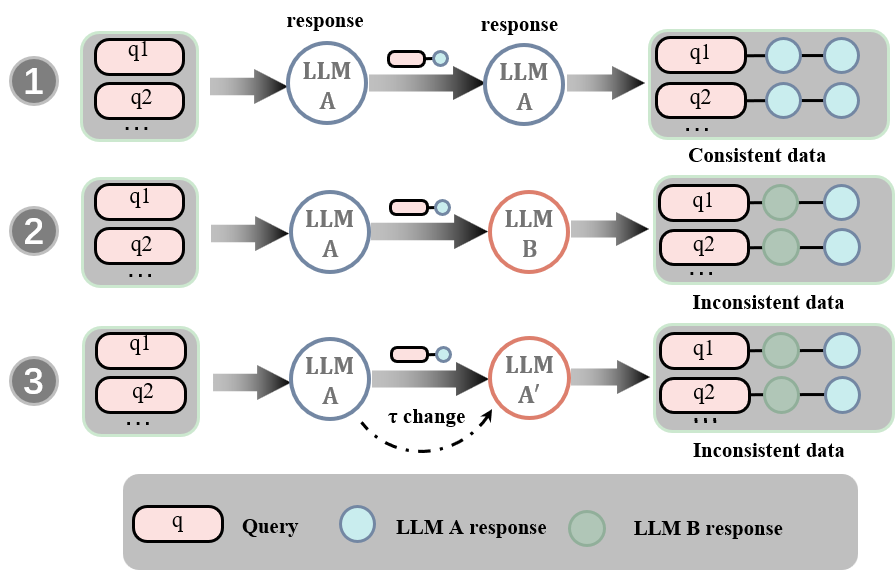}
    \caption{Dataset Construction Rules}  
    \label{data}
    \vspace{-5pt}
\end{figure}

The data craft process is shown in Figure \ref{data}. It includes three methods. In the first method, we input the query into the same large language model twice to obtain consistent response pairs. For inconsistent response pairs, we have two methods: either input the query into two completely different large language models, or input the query into the large language model once, then change the model's temperature coefficient, and input the query into the large language model again.


\section{SimCT Demo}
\begin{figure*}[t]
   \centering
   \includegraphics[width=\linewidth]{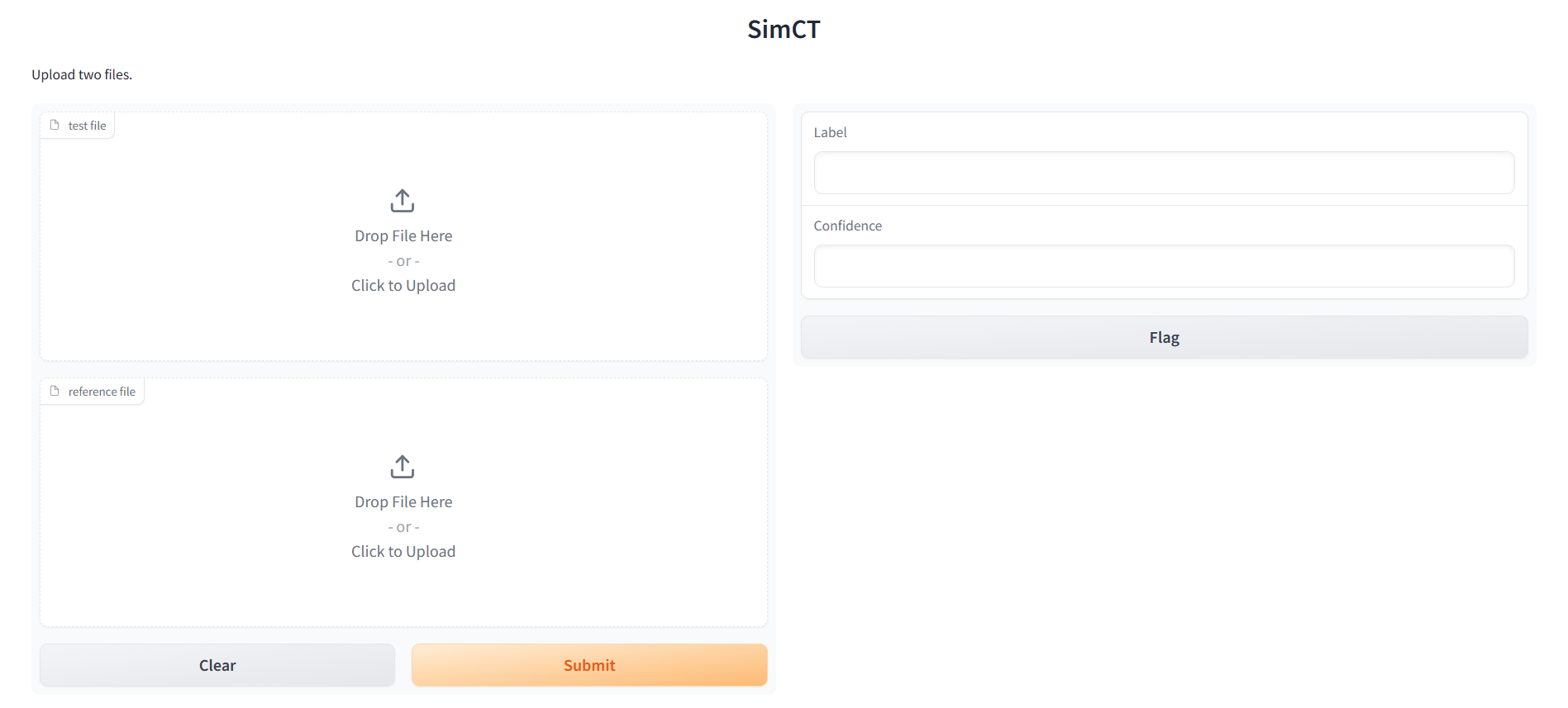}
    \caption{SimCT Demo}  
    \label{demo}
    \vspace{-5pt}
\end{figure*}

\begin{figure*}[t]
   \centering
    \subfloat[Demo for inconsistent model pair test]{
		\includegraphics[width=\linewidth]{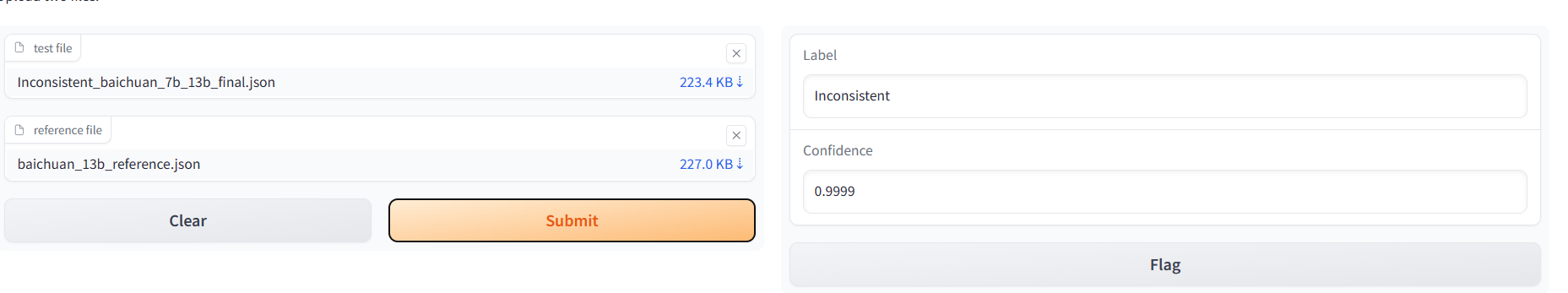}}\\
    \subfloat[Demo for consistent model pair test]{
		\includegraphics[width=\linewidth]{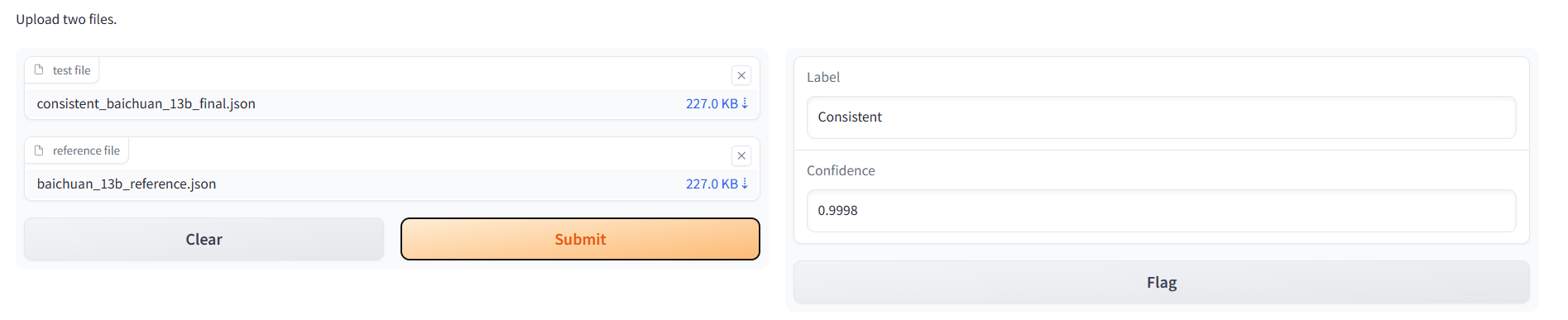}}
    \caption{Illustrative cases for SimCT Demo.}  
    \label{demo case}
    \vspace{-5pt}
\end{figure*}
We design a demo for SimCT using Gradio \citep{abid2019gradio}. The web interface is shown in Figure \ref{demo}. This demo requires users to provide two outputs: one for the test data (to obtain $\mathrm{CT}_{AB}$) and one for the reference data (to obtain $\mathrm{CT}_{AA}$). The final outputs are labels indicating whether two models are consistent and the associated confidence level.

Note that the label and confidence are derived from the P-value of the T-test. Specifically, as hypothesised in main text (i.e., the mean of $\mathrm{CT}_{AB}^{q_i}$ is not equal to the mean of $\mathrm{CT}_{AA}^{q_i}$), we consider model pairs with p-value <= 0.05 to be consistent and those with p-value > 0.05 to be inconsistent. Then for confidence level, it's equal to $1-\mathrm{p-value}$ if label==`consistent' and p-value if label==`inconsistent'.

\end{document}